\documentclass{article} 
\usepackage{iclr2023_conference,times}

\usepackage{amsmath,amsfonts,bm}

\def\eqref#1{equation~\ref{#1}}

\def\1{\bm{1}}

\DeclareMathAlphabet{\mathsfit}{\encodingdefault}{\sfdefault}{m}{sl}
\SetMathAlphabet{\mathsfit}{bold}{\encodingdefault}{\sfdefault}{bx}{n}

\usepackage{hyperref}
\usepackage{url}
\usepackage{graphicx}
\usepackage{subcaption}
\usepackage{amsmath}
\usepackage{algorithm}
\usepackage{algpseudocode}

\usepackage{multirow}
\usepackage{booktabs}
\usepackage{enumerate}
\usepackage{enumitem}
\usepackage[capitalise]{cleveref}
\usepackage{nameref}

\title{Federated Training of Dual Encoding Models on Small Non-IID Client Datasets}

\author{Raviteja Vemulapalli, Warren Richard Morningstar, Philip Andrew Mansfield,\\
\textbf{Hubert Eichner, Karan Singhal, Arash Afkanpour, Bradley Green}\\
Google Research\\
\texttt{\{ravitejavemu,wmorning,memes,huberte,}\\
\texttt{\ karansinghal,arashaf,brg\}@google.com}
}

\iclrfinalcopy 
\begin{document}

\maketitle

\begin{abstract}
Dual encoding models that encode a pair of inputs are widely used for representation learning. Many approaches train dual encoding models by maximizing agreement between pairs of encodings on centralized training data. However, in many scenarios, datasets are inherently decentralized across many clients, motivating federated learning. In this work, we focus on federated training of dual encoding models on decentralized data composed of many small, non-IID (independent and identically distributed) client datasets. Existing approaches require large and diverse training batches to work well and perform poorly when naively adapted to the setting of small, non-IID client datasets using federated averaging. We observe that large-batch loss computation can be simulated on small individual clients for loss functions that are based on encoding statistics. Based on this insight, we propose a novel federated training approach, \textit{Distributed Cross Correlation Optimization (DCCO)}, which trains dual encoding models using encoding statistics aggregated across clients, without sharing individual samples or encodings. Our experimental results on two datasets demonstrate that the proposed approach outperforms federated variants of existing approaches by a large margin.
\end{abstract}

\section{Introduction}
Dual encoding models are a class of models that generate a pair of encodings for a pair of inputs using one or two encoding networks. These models are widely used for representation learning of both unimodal and multi-modal data~\citep{SIMCLR,SimSiam,MOCO,BYOL,BTwins,CLIP, ALIGN}. While several approaches exist for training dual encoding models in centralized settings, training these models on decentralized datasets is less explored.

Federated learning~\citep{FedAvg} is a widely-used approach for learning from decentralized datasets without transferring raw data to a central server. In many real-world scenarios, individual client datasets are small and non-IID (independent and identically distributed), e.g., in \emph{cross-device} federated settings \citep{FedGuide,wang2021field}. For example, in mobile medical apps such as Aysa~\citep{AysaUrl} and DermAssist~\citep{DermAssistUrl}, each user contributes only a few (1-3) images. Motivated by this, we focus on federated training of dual encoding models on decentralized data composed of a large number of small, non-IID client datasets.


Recently, several approaches have been proposed for training dual encoding models in centralized settings based on contrastive losses~\citep{MOCO, SIMCLR, SIMCLR-V2}, statistics-based losses~\citep{BTwins,VICReg}, and predictive losses~\citep{BYOL,SimSiam}. One way to enable federated training of dual encoding models is to adapt these existing approaches using the Federated Averaging (FedAvg) strategy of~\cite{FedAvg}. As described in \cref{sec:problems_with_existing}, all of these approaches require large and diverse training batches to work well, and their performance degrades when trained on small, non-IID client datasets.

To enable large-batch federated contrastive training,~\citet{FURL} and~\citet{FedContrast} propose to share individual sample encodings between clients, raising privacy concerns. ~\citet{FedU, DAFSSL} extend BYOL~\citep{BYOL} to federated settings by using a separate target encoder on each client. ~\citet{SSFL} extends SimSiam~\citep{SimSiam} to federated settings by using a separate personalized model on each client in addition to a shared model.~\citet{HeteroFSSL} goes a step further and removes the shared model. All of these approaches focus on cross-silo settings with small number of clients and thousands of samples per client.~\citet{HeteroFSSL} showed that the performance of these methods degrades significantly as the number of clients increases. Unlike these works, we focus on datasets composed of a large number of small, non-IID client datasets.

In this work we observe that, in the case of statistics-based loss functions, we can simulate large-batch loss computation on each individual (small) client, by first aggregating encoding statistics from many clients and then sharing these aggregated large-batch statistics with all the clients that contributed to them. Based on this observation, we propose a novel approach, \textit{Distributed Cross Correlation Optimization (DCCO)}, for federated training of dual encoding models on small, non-IID client datasets. The proposed approach simulates large-batch training with the loss function of~\citet{BTwins}, which we refer to as \textit{Cross Correlation Optimization (CCO)} loss. This is achieved without sharing individual data samples or their encodings between clients. 

Federated dual encoding models have also been explored for recommendation systems~\citep{FederatedDualEncoder, fedcl,augenstein2022mixed}, where a lookup table of item encodings is used to train on non-IID data. Here we focus on continuous signals such as images without a lookup table.\vspace{-5pt}

\subsection*{Major Contributions}
\begin{itemize}[topsep=0pt,itemsep=-0.5ex,partopsep=0ex,parsep=1ex,leftmargin=5ex] 
    \item We observe that large-batch training of dual encoding models can be simulated on decentralized datasets by using loss functions based on encoding statistics aggregated across clients, without sharing individual samples or their encodings.
    \item Building on this insight, we present Distributed Cross Correlation Optimization (DCCO), a novel approach for training dual encoding models on decentralized datasets composed of a large number of small, non-IID client datasets. 
    \item We prove that when we perform one step of local training in each federated training round, one round of DCCO training is equivalent to one step of centralized training on a large batch composed of all samples across all clients participating in the federated round.
    \item We evaluate the proposed approach on two datasets and show that it outperforms FedAvg variants of contrastive and CCO training by a significant margin. The proposed method also significantly outperforms supervised training from scratch, demonstrating its effectiveness for decentralized self-supervised learning.
\end{itemize}

\section{Problems with Existing Approaches}
\label{sec:problems_with_existing}
\paragraph{Contrastive loss functions} Contrastive losses maximize the similarity between two encodings of a data sample while pushing encodings of different samples apart. This is highly effective when each sample is contrasted against a large set of diverse samples~\citep{SIMCLR,SIMCLR-V2,MOCO}.  When combined with FedAvg, the effectiveness of contrastive training decreases as each sample is contrasted against a small set of within-client samples, which may be relatively similar.

\paragraph{Dependence on batch normalization}
Approaches such as BYOL~\citep{BYOL} and SimSiam~\citep{SimSiam} use predictive losses that encourage two encodings of a data sample to be predictive of each other. Though these approaches do not explicitly push the encodings of different samples apart, they work well when trained with large batches in centralized settings. Importantly, they use batch normalization~\citep{BatchNorm}, whose efficacy decreases rapidly when batches become smaller~\citep{GroupNorm}. When training on small, non-IID client datasets in federated settings, group normalization~\citep{GroupNorm} is typically used instead of batch normalization~\citep{FLGroupNorm,FedVisClass}. When we experimented with BYOL and SimSiam by replacing batch normalization with group normalization, the models did not train well (see \cref{sec:batchnorm_ablations}). This dependence on batch normalization suggests that these approaches are not a good fit for federated training on small, non-IID client datasets.

\paragraph{Statistics-based loss functions} While the above contrastive and predictive losses directly use individual sample encodings in their computation, the CCO loss introduced by \citet{BTwins} is a function of encoding statistics computed over a batch of samples. This loss function maximizes the correlation coefficient values of matching dimensions and minimizes the correlation coefficient values of non-matching dimensions of the two encodings of a dual encoding model. Since CCO loss is a function of batch statistics, its efficacy decreases with smaller batch sizes, and it performs poorly in federated settings when used for training on small, non-IID client datasets (see Sec.~\ref{sec:experiments}).

\section{Proposed Approach}
\subsection{Cross Correlation Optimization Loss}
Let $X$ and $Y$ be the two inputs to a dual encoding model, and $F = [F_i]\in \mathcal{R}^d$ and $G = [G_j] \in \mathcal{R}^d$ be their encodings, respectively. The CCO loss used for training dual encoding models in~\citet{BTwins} is given by~\footnote{~\citet{BTwins} do not refer to this loss as CCO loss; we use this more general term instead of \textit{BarlowTwins} loss to emphasize the loss function's applicability to cases where the two encoder networks differ.} 
\begin{equation}
\mathcal{L}_{CCO} = \sum_{i=1}^d (1 - C_{ii})^2 + \lambda \sum_{i=1}^d \frac{1}{d-1}\sum_{\substack{j=1\\j\neq i}}^d C_{ij}^2,\ \ C_{ij} = \frac{\left<F_iG_j\right> - \left<F_i\right> \left<G_j\right>}{\sqrt{\left<(F_i)^2\right> - \left<F_i\right>^2}\sqrt{\left<(G_j)^2\right> - \left<G_j\right>^2}},
\label{eq:cco_loss}
\end{equation}
where $C_{ij}$ represents the correlation coefficient between $i^{th}$ component of $F$ and $j^{th}$ component of $G$. Here, $\left<\ \right>$ represents average values computed using a batch of samples. The first term in the CCO loss encourages the two encodings of a data sample to be similar by maximizing the correlation coefficient between the matching dimensions of the encodings, and the second term reduces the redundancy between output units by decorrelating the different dimensions of the encodings.

\subsection{Motivation: Aggregating and Redistributing Encoding Statistics}
Both contrastive and CCO losses are highly effective when used with large training batches, and their efficacy decreases as batch size decreases. Contrastive losses directly use individual sample encodings, and it is unclear how large-batch contrastive loss can be computed on small clients without sharing individual sample encodings between clients. Unlike contrastive losses, CCO loss only uses encoding statistics $\left<F_i\right>, \left<(F_i)^2\right>, \left<G_j\right>, \left<(G_j)^2\right>, \left<F_iG_j\right>$. Although CCO loss is a nonlinear function of these statistics, the statistics themselves are average values, i.e. linear combinations of individual sample values. Hence, we can compute large-batch statistics as a weighted average of small-batch statistics computed on individual clients, without sharing individual sample encodings. This makes it possible to simulate large-batch CCO loss on individual clients by first aggregating encoding statistics from many clients and sharing these aggregated large-batch statistics with all the clients that contributed to them. Based on this insight, we propose our DCCO approach that simulates large-batch training using small, non-IID client datasets.

The strategy of aggregating and redistributing statistics  can also be used to implement multi-client synchronous batch normalization in federated settings. However, since batch normalization is typically used at every layer of the encoder network, this will add multiple additional rounds of server-client communication (as many as the number of batch normalization layers in the network) which is undesirable. In contrast, the proposed DCCO approach uses only one additional round of communication by aggregating statistics at the end of the network before computing the loss function.

\subsection{Distributed Cross Correlation Optimization (DCCO)}
\begin{figure}
\begin{center}
\includegraphics[clip, trim=0.1cm 3cm 0.1cm 4.3cm, width=0.95\textwidth]{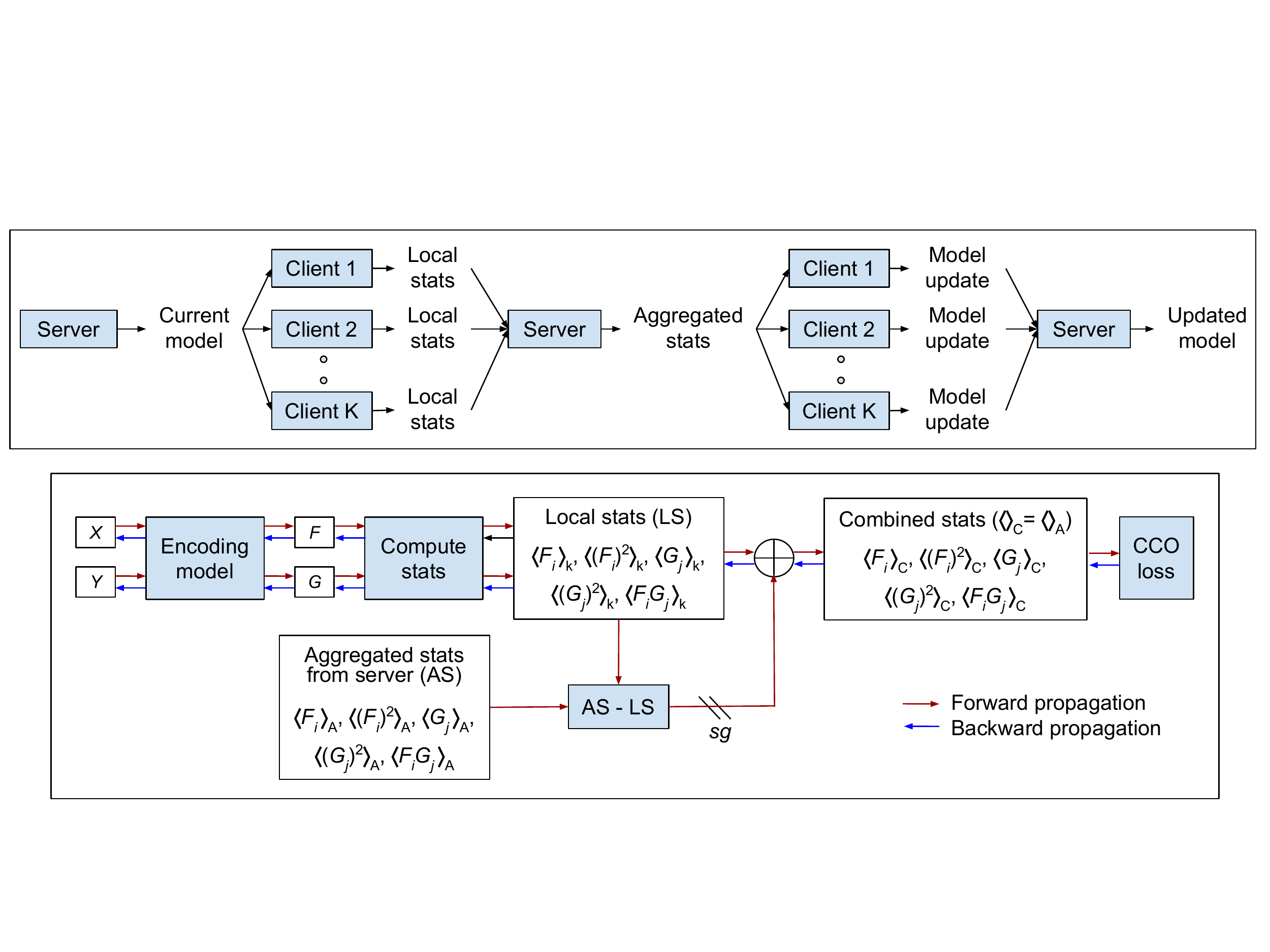}
\end{center}
\caption{Top: one round of DCCO training. Bottom: local training on client $k$. Here \textit{sg} refers to the stop gradient operation, and $\bigoplus$ denotes addition. The combined statistics $\left<\ \right>_C$ used for loss computation are equal to the aggregated statistics $\left<\ \right>_A$ from the server. Note that gradient backpropagation happens only through the local statistics $\left<\ \right>_k$.}
\label{fig:DCCO}
\end{figure}
Figure~\ref{fig:DCCO} presents the proposed DCCO training approach. In each training round, a central server samples $K$ clients from the pool of available clients and broadcasts the current model to these clients. Each client $k$ uses this model to encode its local data and compute local encoding statistics $\left<F_i\right>_k, \left<(F_i)^2\right>_k, \left<G_j\right>_k, \left<(G_j)^2\right>_k, \left<F_iG_j\right>_k$ based on $N_k$ samples. These local encoding statistics are then aggregated by the central server using weighted averaging, 
and the aggregated statistics are shared with the participating clients. During this aggregation, each local encoding statistic is given a weight that is proportional to the number of samples that contributed to its computation.
Then, each client computes a local model update by minimizing the CCO loss function computed using aggregated statistics $\left<\ \right>_A$. During local training, while aggregated statistics are used for computing the loss function, gradients are backpropagated only through local statistics $\left<\ \right>_k$ since each client has access only to its local data. Finally, the local model updates are aggregated by the central server and used to update the current model. To aggregate client model updates, we use a weighted average, weighted by examples per client. During the entire training process, individual data samples never leave the clients either in raw or encoded formats. 

Our statistics aggregation procedure is same as the FedAvg weight aggregation procedure and thus may contain similar vulnerabilities (from a privacy standpoint) which can be addressed with approaches such as Secure Aggregation~\citep{SecureAgg}. We plan to explore this in the future.

Our work focuses on training with small client datasets (for example, no more than six images per client in the case of DERM dataset). Hence, in each federated training round, we only perform one step of local training on each participating client. In this setting, one round of DCCO training is equivalent to one centralized training step on a batch composed of all data samples across all clients participating in the federated round. See \cref{sec:Appendix_proof} for the proof. This equivalence between federated and centralized training holds only because of the additional step of aggregating and sharing encoding statistics. Naively applying FedAvg with one within-client CCO loss-based training step per round is not equivalent to centralized training, as we show in \cref{sec:experiments}.

While this work focuses on the CCO loss of~\citet{BTwins}, the proposed distributed learning strategy can also be used with other statistics-based loss functions such as~\cite{VICReg}.

\section{Experiments}
\label{sec:experiments}
In this section, we evaluate the proposed DCCO approach in the context of self-supervised learning where the two inputs of a dual encoding model are generated by applying random augmentations to a single unlabeled input, and both inputs are processed using the same network.

\paragraph{Experimental setup} First, the encoder network is pretrained on an unlabeled, decentralized dataset. Then, a linear classifier is added on top of the encoder network, and either only the newly added classifier (linear evaluation protocol) or the entire network (full finetuning protocol) is trained using a small labeled classification dataset on a central sever. The accuracy of final model is used as the evaluation metric.

\paragraph{Comparisons} We compare the proposed approach with FedAvg variants of within-client CCO and contrastive loss-based training. We use CCO loss-based centralized training as upper bound for the proposed approach. We report results for fully-supervised training from scratch (using the limited labeled data) to demonstrate the effectiveness of DCCO as a self-supervised pretraining strategy.

\paragraph{CIFAR-100 dataset~\citep{cifar}} We use all 50K training images in the dataset as unlabeled data for pretraining, and a small fraction of them (10\% or 1\%) as labeled data for supervised finetuning. To study the effect of non-identical client data distribution, we generated IID and non-IID client datasets using the Dirichlet distribution-based sampling process of~\cite{HsuQiBrown2019} with $\alpha = 1000$ and $\alpha = 0$, respectively. We generated multiple decentralized datasets by varying the total number of clients and the number of samples per client (see \cref{tab:cifar_results}).

\paragraph{DERM dataset~\citep{Dermdata}}
This dataset consists of de-identified images of skin conditions captured using consumer-grade digital cameras. Each case in the dataset includes one to six images of a skin condition from a single patient. This naturally partitioned dataset enables measuring generalization to patients unseen during training \citep{yuan2021we}. A portion of this dataset is labeled with 419 skin conditions. Following~\cite{REMEDIS}, we focus on the most common 26 conditions and group the rest into `Other` class leading to a label space of 27 categories. As in~\citet{REMEDIS}, we use 207,032 unlabeled images (59,770 cases) for pretraining the encoder network. For federated pretraining, we consider each case as a separate client. For supervised finetuning, we use up to 4,592 labeled cases. The final accuracy is reported on a test set of 4,146 labeled cases. This dataset also provides a validation split with 1,190 labeled cases which we use for tuning.

\paragraph{Networks} Residual networks~\citep{ResNet} with 14 and 50 layers are used as encoder networks for experimenting with CIFAR-100 and DERM datasets, respectively. Following~\cite{BTwins}, a three layer fully-connected projection network is used to increase the dimensionality of the encodings before computing the CCO loss during pretraining, and the projection network is discarded while training the final classifier. The projection network configurations are [1024, 1024, 1024] and [2048, 2048, 4096] for CIFAR-100 and DERM datasets, respectively. For the contrastive loss, following~\cite{SIMCLR, SIMCLR-V2}, a fully-connected projection network is used to reduce the dimensionality of the encodings before computing the loss. The projection network configurations are [256, 256, 128] and [2048, 2048, 128] for CIFAR-100 and DERM datasets, respectively. Weight standardization~\citep{WeightStandard} and group normalization~\citep{GroupNorm} with 32 groups are used at every layer except the last projection layer. For the DERM dataset, since each case consists of multiple images (up to six), the final classification network performs average pooling of individual image encodings and uses a linear classifier on top of the average-pooled feature to predict the label for a case. Following~\cite{DermSSL}, we use $224 \times 224$ images as input while pretraining the encoder, and $448 \times 448$ images as input while training the classifier.

\subsection{Training details}
\paragraph{CIFAR-100} We experimented with several decentralized versions by varying the total number of clients and the number of samples per client (see Table~\ref{tab:cifar_results}). During federated pretraining, all models were trained for 100K rounds. For FedAvg variants of CCO and contrastive loss-based training, we observed overfitting. For these approaches, we evaluated multiple pretrained checkpoints under linear evaluation protocol and report the results for the best checkpoints. We did not observe such overfitting for the proposed DCCO approach and report results based on 100K rounds of training. Each client is visited around 1000 times during federated pretraining of 100K rounds. So, for centralized CCO loss-based pretraining, we used 1000 epochs of training with a batch size of 512.

\paragraph{DERM} We varied the number of clients sampled in each round (see Table~\ref{tab:derm_resutls}). All models were trained for 75K rounds during federated pretraining. For this dataset (59,770 clients), each client is visited around 320 times if we perform 75K rounds of federated training with 256 clients per round. So, for centralized CCO training, we used 320 epochs with a batch size of 512.

We used a value of 20 for the tradeoff parameter $\lambda$ in CCO loss (Eq.~\ref{eq:cco_loss}) and a value of 0.1 for the temperature parameter in contrastive loss~\citep{SIMCLR}. In all federated experiments, we used gradient descent with learning rate 1.0 as the local optimizer on clients. Please refer to \cref{sec:appendix_hyperparams} for further details about the training hyperparameters and data augmentations used in all settings.

\subsection{Results}
\paragraph{CIFAR100}
Table~\ref{tab:cifar_results} shows the performance of various approaches under linear evaluation and full finetuning protocols for multiple decentralized versions of CIFAR-100 dataset.  When only 500 images are labeled, training only the linear classifier performed better than finetuning the entire network. So, in this case, we only report results under linear evaluation protocol. The proposed approach outperforms the FedAvg variants of contrastive and CCO training by a significant margin for both IID and non-IID client datasets. Specifically, in the case of non-IID client datasets, the performance gains are in the range of 11-18\% under linear evaluation protocol and 6-10\% under full finetuning protocol, and in the case of IID client datasets, the gains are in the range of 4-10\% under linear evaluation protocol and 4-6\% under full finetuning protocol. The proposed approach also outperforms fully-supervised training from scratch by 10-20\%, demonstrating the effectiveness of DCCO as a federated self-supervised pretraining strategy.
\begin{table}[t]
    \begin{small}
    \begin{center}
    \begin{sc}
    \begin{tabular}{lcccc|ccc}
    \toprule
    \multicolumn{1}{c}{} & \multicolumn{4}{c|}{Non-IID client datasets} & \multicolumn{3}{|c}{IID client datasets}\\\hline
     Total clients &  50,000 & 12,500 & 6,250 & 3,125 & 12,500 & 6,250 & 3,125\\ \hline
    Samples / Client & 1  & 4 & 8 & 16 & 4 & 8 & 16\\ \hline
    Clients / Round & 512 & 128 & 64 & 32 & 128 & 64 & 32\\ 
    \bottomrule
    \multicolumn{8}{c}{Linear evaluation protocol (5K labeled training samples)}\\
    \hline
     CCO + FedAvg & --  & Failed & 28.7 & 30.7 & Failed &  41.2 & 44.1\\\hline
    Contrastive + FedAvg & -- & 31.1 & 32.4 & 31.8 & 41.1 & 44.1 & 46.4\\\hline
    Proposed DCCO & 51.7 & 49.6 & 48.1 & 45.9 & 51.4 & 52.0 & 51.9\\\hline
    Centralized CCO & \multicolumn{7}{c}{52.6} \\\hline
    Supervised from scratch & \multicolumn{7}{c}{42.4} \\\bottomrule
    \multicolumn{8}{c}{Full finetuning protocol (5K labeled training samples)}\\\hline
    CCO + FedAvg & --  & Failed & 36.3 & 38.9 & Failed &  43.0 & 46.0\\\hline
    Contrastive + FedAvg & -- & 40.8 & 41.8 & 41.3 & 44.9 & 46.8 & 47.4\\\hline
    Proposed DCCO & 51.5 & 50.5 & 49.3 & 47.7 & 51.5 & 52.1 & 52.0\\\hline
    Centralized CCO & \multicolumn{7}{c}{52.5} \\\hline
    Supervised from scratch & \multicolumn{7}{c}{42.4} \\\bottomrule    
    \multicolumn{8}{c}{Linear evaluation protocol (500 labeled training samples)}\\\hline
    CCO + FedAvg & -- & Failed & 13.5 & 14.5 & Failed & 26.2 & 27.6\\\hline
    Contrastive + FedAvg & -- & 14.8 & 15.6 & 15.0 & 24.7 & 27.8 & 29.3\\\hline
    Proposed DCCO & 33.3 & 31.2 & 29.5 & 26.7 & 33.3 & 33.5 & 33.5\\\hline
    Centralized CCO & \multicolumn{7}{c}{34.0} \\\hline
    Supervised from scratch & \multicolumn{7}{c}{13.1} \\\bottomrule
    \end{tabular}
    \end{sc}
    \end{center}
    \end{small}
    \caption{Results on CIFAR-100 dataset under linear evaluation and full finetuning protocols.}
    \label{tab:cifar_results}
\end{table}

In the case of non-IID client datasets, for a fixed global batch size (i.e., the total number of samples participating in a round), the performance of DCCO approach increases as the number of samples per client decreases and the number of clients per round increases, approaching the performance of centralized CCO training when each client has only one sample~\footnote{Ideally, when each client has only one sample, the performance of DCCO with 512 clients per round should be close to the performance of centralized CCO with a batch size of 512. There is a small ($\sim$1\%) performance gap between the two due to the differences in the random data augmentation pipelines implemented in Tensorflow (used for centralized training) and Tensorflow Federated (used for federated training).}. This is because the proposed DCCO approach optimizes a loss function based on aggregated global batch statistics and using more clients increases the diversity of samples in the global batch. However, we do not see such trend in the case of IID client datasets because each sample in every client dataset is already sampled in an IID fashion from the full CIFAR-100 dataset. 

We do not report results for FedAvg variants of contrastive and CCO losses when each client has only one sample because we need at least two samples to compute these loss functions. The proposed DCCO approach can still be used in this case since it uses statistics aggregated from multiple clients to compute the loss. We do not report results for FedAvg variant of CCO loss when each client has only four samples because training was unstable in this case.

\paragraph{DERM}
Table~\ref{tab:derm_resutls} shows the performance of various approaches under full finetuning protocol for different amounts of labeled data. The proposed approach clearly outperforms the FedAvg variant of contrastive loss-based training and achieves a performance close to that of centralized CCO training. The proposed approach also results in significant performance gains (13-17\%) when compared to fully-supervised training from scratch demonstrating its effectiveness in leveraging unlabeled decentralized datasets. We do not report results for FedAvg variant of CCO loss because training was unstable in this case as each client in this dataset has a maximum of six images.

\begin{table}[t]
    \begin{small}
    \begin{center}
    \begin{sc}
    \begin{tabular}{l|c|c|c|c|c|c|c|c|c}
    \toprule
        \multicolumn{1}{l}{Labeled data (Finetuning)} & \multicolumn{3}{|c|}{1,524 cases} & \multicolumn{3}{|c|}{3,057 cases} & \multicolumn{3}{|c}{4,592 cases}\\\hline
        Num. clients (Pretraining) &  \multicolumn{3}{c|}{59,770} & \multicolumn{3}{c|}{59,770} & \multicolumn{3}{c}{59,770}\\\hline
         Clients / Round & 64 & 128 & 256 & 64 & 128 & 256 &  64 & 128 & 256\\
         \bottomrule
         Contrastive + FedAvg & 38.9 & 41.8 & 43.9 & 45.3 & 46.7 & 47.8 & 47.7 & 49.1 & 51.5\\
         \hline
         Proposed DCCO & 42.4 & 46.3 & 47.8 & 46.7 & 50.3 & 51.1 & 48.3 & 52.8 & 52.5\\
         \hline
         Centralized CCO & \multicolumn{3}{c|}{48.0} & \multicolumn{3}{c|}{52.0} & \multicolumn{3}{c}{53.6}\\
         \hline
         Supervised from scratch & \multicolumn{3}{c|}{30.6} & \multicolumn{3}{c|}{35.9} & \multicolumn{3}{c}{39.2}\\
         \bottomrule
    \end{tabular}
    \end{sc}
    \end{center}
    \end{small}
    \caption{Results on DERM dataset under full finetuning protocol.}
    \label{tab:derm_resutls}
\end{table}

\section{Conclusion and Future Work}
In this work, we proposed an approach for training dual encoding models on decentralized datasets composed of a large number of small, non-IID client datasets. The proposed approach optimizes a loss function based on encoding statistics and simulates large-batch loss computation on individual clients by using encoding statistics aggregated across multiple clients. When each client participating in a training round performs only one local training step, each federated round of DCCO training is equivalent to a centralized training step on a batch consisting of all data samples participating in the corresponding round. Our experimental results on two image datasets show that the proposed approach outperforms FedAvg variants of within-client contrastive and CCO loss-based training. Our proposed approach also outperforms supervised training from scratch by a significant margin; demonstrating its effectiveness as a federated self-supervised learning approach. 

This paper focused on unimodal datasets and the CCO loss function of~\citet{BTwins}. In the future, we plan to experiment with multi-modal datasets, and other statistics-based loss functions such as~\citet{VICReg}. There are also several other interesting research directions we plan to pursue in the future. Extending the proposed approach to large client datasets is an interesting direction. When training on large client datasets, we may need to perform multiple steps of local training on each client. In each step, only a small subset of samples that are contributing to the loss function participate in gradient computation. Also, while the model weights change after each local step, the aggregated statistics used in the loss function remain constant. Addressing the effects of these \textit{partial gradients} and \textit{stale statistics} is another challenge. The proposed DCCO approach uses two communication rounds between server and clients within one federated training round, and reducing this to one round, e.g., by using statistics from a previous round, is promising.

\bibliography{iclr2023_conference}
\bibliographystyle{iclr2023_conference}

\appendix
\section{Proof of Equivalence to Centralized Training}
\label{sec:Appendix_proof}
\textbf{Claim:} When we perform one step of local training on each participating client in a federated training round, one round of DCCO training is equivalent to one centralized training step on a batch composed of all data samples participating in the round.\\[10pt]
\textbf{Proof:}  Let $F_i^n$ and $G_j^n$ respectively denote the $i^{th}$ and $j^{th}$ components of encodings $F$ and $G$ of $n^{th}$ sample on a client. Based on the definition of $\left<F_i\right>_C$, we get 
\begin{equation}
\left<F_i\right>_C = \left<F_i\right>_k + StopGradient\left[\left<F_i\right>_A - \left<F_i\right>_k\right] \implies \frac{\partial \left<F_i\right>_C}{\partial F_i^n} = \frac{\partial \left<F_i\right>_k}{\partial F_i^n}.
\label{eqn:sg_effect1}
\end{equation} 
Similarly, by definitions of  $\left<(F_i)^2\right>_C, \left<G_j\right>_C,  \left<(G_j)^2\right>_C,  \left<F_iG_j\right>_C$, we get
\begin{align}
 \frac{\partial \left<(F_i)^2\right>_C}{\partial F_i^n} = \frac{\partial \left<(F_i)^2\right>_k}{\partial F_i^n}, \frac{\partial \left<G_j\right>_C}{\partial G_j^n} &= \frac{\partial \left<G_j\right>_k}{\partial G_j^n}, \frac{\partial \left<(G_j)^2\right>_C}{\partial G_j^n} = \frac{\partial \left<(G_j)^2\right>_k}{\partial G_j^n},\nonumber\\ 
 \frac{\partial \left<F_iG_j\right>_C}{\partial F_i^n} = \frac{\partial \left<F_iG_j\right>_k}{\partial F_i^n}&, \frac{\partial \left<F_iG_j\right>_C}{\partial G_j^n} = \frac{\partial \left<F_iG_j\right>_k}{\partial G_j^n}
 \label{eqn:sg_effect2}
\end{align} 
By chain rule, we have
\begin{align}
\frac{\partial\mathcal{L}_{CCO}}{\partial F^n_i} &= \frac{\partial\mathcal{L}_{CCO}}{\partial \left<F_i\right>_C}\frac{\partial \left<F_i\right>_C}{\partial F^n_i} + \frac{\partial\mathcal{L}_{CCO}}{\partial \left<(F_i)^2\right>_C}\frac{\partial \left<(F_i)^2\right>_C}{\partial F^n_i} + \sum_{j=1}^d \frac{\partial\mathcal{L}_{CCO}}{\partial \left<F_iG_j\right>_C}\frac{\partial \left<F_iG_j\right>_C}{\partial F^n_i} \nonumber\\
\frac{\partial\mathcal{L}_{CCO}}{\partial G^n_j} &= \frac{\partial\mathcal{L}_{CCO}}{\partial \left<G_j\right>_C}\frac{\partial \left<G_j\right>_C}{\partial G^n_j} + \frac{\partial\mathcal{L}_{CCO}}{\partial \left<(G_j)^2\right>_C}\frac{\partial \left<(G_j)^2\right>_C}{\partial G^n_j} + \sum_{i=1}^d \frac{\partial\mathcal{L}_{CCO}}{\partial \left<F_iG_j\right>_C}\frac{\partial \left<F_iG_j\right>_C}{\partial G^n_j}
\label{eqn:chain}
\end{align}
Substituting Eq.~\ref{eqn:sg_effect1} and~\ref{eqn:sg_effect2} in Eq.~\ref{eqn:chain}, we get 
\begin{align}
\frac{\partial\mathcal{L}_{CCO}}{\partial F^n_i} &=
\frac{\partial\mathcal{L}_{CCO}}{\partial \left<F_i\right>_C}\frac{\partial \left<F_i\right>_k}{\partial F^n_i} + \frac{\partial\mathcal{L}_{CCO}}{\partial \left<(F_i)^2\right>_C}\frac{\partial \left<(F_i)^2\right>_k}{\partial F^n_i} + \sum_{j=1}^d \frac{\partial\mathcal{L}_{CCO}}{\partial \left<F_iG_j\right>_C}\frac{\partial \left<F_iG_j\right>_k}{\partial F^n_i} \nonumber \\
&= \frac{1}{N_k}\frac{\partial\mathcal{L}_{CCO}}{\partial \left<F_i\right>_C} + \frac{2}{N_k}\frac{\partial\mathcal{L}_{CCO}}{\partial \left<(F_i)^2\right>_C} F^n_i + \frac{1}{N_k}\sum_{j=1}^d \frac{\partial\mathcal{L}_{CCO}}{\partial \left<F_iG_j\right>_C}G^n_j \nonumber \\
\frac{\partial\mathcal{L}_{CCO}}{\partial G^n_j} &=
\frac{\partial\mathcal{L}_{CCO}}{\partial \left<G_j\right>_C}\frac{\partial \left<G_j\right>_k}{\partial G^n_j} + \frac{\partial\mathcal{L}_{CCO}}{\partial \left<(G_j)^2\right>_C}\frac{\partial \left<(G_j)^2\right>_k}{\partial G^n_j} + \sum_{i=1}^d \frac{\partial\mathcal{L}_{CCO}}{\partial \left<F_iG_j\right>_C}\frac{\partial \left<F_iG_j\right>_k}{\partial G^n_j}\nonumber  \\
&= \frac{1}{N_k}\frac{\partial\mathcal{L}_{CCO}}{\partial \left<G_j\right>_C} + \frac{2}{N_k}\frac{\partial\mathcal{L}_{CCO}}{\partial \left<(G_j)^2\right>_C} G^n_j + \frac{1}{N_k}\sum_{i=1}^d \frac{\partial\mathcal{L}_{CCO}}{\partial \left<F_iG_j\right>_C}F^n_i
\label{eqn:chain_sg_effect}
\end{align}
By chain rule, the gradients for encoding model parameters $\theta$ on $k^{th}$ client are given by
\begin{align}
\frac{\partial\mathcal{L}_{CCO}}{\partial \theta} \bigg|_k &= \sum_{n=1}^{N_k} \sum_{i=1}^d\frac{\partial\mathcal{L}_{CCO}}{\partial F^n_i}\frac{\partial F^n_i}{\partial \theta} + \sum_{n=1}^{N_k} \sum_{j=1}^d\frac{\partial\mathcal{L}_{CCO}}{\partial G^n_j}\frac{\partial G^n_j}{\partial \theta},
\label{eqn:grad}
\end{align}
where $N_k$ is the number of samples on the client that contributed to $\mathcal{L}_{CCO}$.

Substituting Eq.~\ref{eqn:chain_sg_effect} in Eq.~\ref{eqn:grad}, we get
\begin{align}
\frac{\partial\mathcal{L}_{CCO}}{\partial \theta}\bigg|_k =
&\frac{1}{N_k}\sum_{n=1}^{N_k} \sum_{i=1}^d \left(\frac{\partial\mathcal{L}_{CCO}}{\partial \left<F_i\right>_C} + 2\frac{\partial\mathcal{L}_{CCO}}{\partial \left<(F_i)^2\right>_C} F^n_i + \sum_{j=1}^d \frac{\partial\mathcal{L}_{CCO}}{\partial \left<F_iG_j\right>_C}G^n_j \right)\frac{\partial F^n_i}{\partial \theta}\hspace{5pt} + \nonumber \\ 
&\frac{1}{N_k}\sum_{n=1}^{N_k} \sum_{j=1}^d
 \left(\frac{\partial\mathcal{L}_{CCO}}{\partial \left<G_j\right>_C} + 2\frac{\partial\mathcal{L}_{CCO}}{\partial \left<(G_j)^2\right>_C} G^n_j + \sum_{i=1}^d \frac{\partial\mathcal{L}_{CCO}}{\partial \left<F_iG_j\right>_C}F^n_i \right)\frac{\partial G^n_j}{\partial \theta}
\label{eqn:client_grad}
\end{align}
The values of stats $\left\{\left<F_i\right>_C\right\}_{i=1}^d$, $\left\{\left<(F_i)^2\right>_C\right\}_{i=1}^d$, $\left\{\left<G_j\right>_C\right\}_{j=1}^d$, $\left\{\left<(G_j)^2\right>_C\right\}_{j=1}^d$, $\left\{\left<F_iG_j\right>_C\right\}_{i,j=1}^{d}$ and loss $\mathcal{L}_{CCO}$ are same on all the clients participating in a DCCO training round. So, when each client performs one step of local training and the server averages the model updates from the clients by weighing them according to the number of contributing samples on each client, the equivalent global model gradient update is given by
\begin{align}
\frac{\partial\mathcal{L}_{CCO}}{\partial \theta} &= \frac{1}{N}\sum_{k=1}^K N_k \frac{\partial\mathcal{L}_{CCO}}{\partial \theta}\bigg|_k \nonumber\\
=&\frac{1}{N}\sum_{n=1}^{N} \sum_{i=1}^d \left(\frac{\partial\mathcal{L}_{CCO}}{\partial \left<F_i\right>_C} + 2\frac{\partial\mathcal{L}_{CCO}}{\partial \left<(F_i)^2\right>_C} F^n_i + \sum_{j=1}^d \frac{\partial\mathcal{L}_{CCO}}{\partial \left<F_iG_j\right>_C}G^n_j \right)\frac{\partial F^n_i}{\partial \theta} \hspace{5pt} + \nonumber \\ 
&\frac{1}{N}\sum_{n=1}^{N} \sum_{j=1}^d
 \left(\frac{\partial\mathcal{L}_{CCO}}{\partial \left<G_j\right>_C} + 2\frac{\partial\mathcal{L}_{CCO}}{\partial \left<(G_j)^2\right>_C} G^n_j + \sum_{i=1}^d \frac{\partial\mathcal{L}_{CCO}}{\partial \left<F_iG_j\right>_C}F^n_i \right)\frac{\partial G^n_j}{\partial \theta}
\label{eqn:global_grad}
\end{align}
where $N$ is the total number of samples participating in the round. 

If all the $N$ samples participating in a DCCO round are present on a single client, then according to Eq.~\ref{eqn:client_grad}, the gradients computed on that client using these $N$ samples will be same as the gradients in Eq.~\ref{eqn:global_grad}. Hence one round of federated DCCO training is equivalent to one step of centralized training on a batch composed of all $N$ samples participating in the federated round.

\section{Training Details}
\label{sec:appendix_hyperparams}
We used a value of 20 for the tradeoff parameter $\lambda$ in CCO loss (Eq.~\ref{eq:cco_loss}) and a value of 0.1 for the temperature parameter in contrastive loss~\citep{SIMCLR}. In all federated experiments, we used gradient descent with learning rate 1.0 as the local optimizer on clients.

\subsection{CIFAR-100}
\textbf{Federated pretraining:} For updating the model on server, we used Adam optimizer~\citep{Adam} with cosine learning rate decay. We experimented with initial learning rates of $5e^{-3}$ and $1e^{-3}$ and report the best results.

\textbf{Centralized pretraining:}We trained for 1000 epochs with a batch size of 512. We used Adam optimizer with initial learning rate of $5e^{-3}$, and cosine learning rate decay.

\textbf{Full network finetuning with 5K samples:} We trained for 100 steps using Adam optimizer with batch size 256, initial learning rate $5e^{-3}$, and cosine learning rate decay.

\textbf{Supervised training from scratch:} We trained for 100 epochs using Adam optimizer with batch size 256, initial learning rate 0.01, and cosine learning rate decay.

\textbf{Linear classifier training with 5K samples:} We trained for 1000 steps using LARS optimizer~\citep{Lars} with batch size 512, initial learning rate of 2.0, cosine learning rate decay, and a momentum of 0.9.

\textbf{Linear classifier training with 500 samples:} We trained for 400 steps using LARS optimizer with batch size 128, initial learning rate of 2.5, cosine learning rate decay, and a momentum of 0.9.

\textbf{Augmentations:} During self-supervised pretraining, we used all the data augmentations from~\citet{BYOL} except blur, and during supervised finetuning we only used flip augmentation. 

\subsection{DERM}
\textbf{Federated pretraining:} For updating the model on server, we used LARS optimizer with cosine learning rate decay and a momentum of 0.9. For DCCO pretraining, the initial learning rate was set to 0.15, 0.9 and 1.8 while using 64, 128 and 256 clients per round, respectively. For FedAvg training with contrastive loss, the initial learning rate was set to 0.15, 0.3 and 0.6 while using 64, 128 and 256 clients per round, respectively. All models were trained for 75K rounds.

\textbf{Centralized pretraining:} We trained for 320 epochs with a batch size of 512. We used LARS optimizer with initial learning rate of 0.6, cosine learning rate decay, and a momentum of 0.9.

\textbf{Supervised training from scratch and full network finetuning:} We used Adam optimizer with a batch size of 128. For each experiment, we used the validation split of this dataset to search for the best learning rate (among $3e^{-4}$ and $1e^{-4}$) and number of training steps (up to 100k). 

\textbf{Augmentations:} During self-supervised pretraining, we used random rotation and all the data augmentations from~\citet{BYOL} except solarization. During supervised finetuning, we used all the data augmentations from~\citet{DermSSL}.

\section{BYOL and SimSiam with Group Normalization}
\label{sec:batchnorm_ablations}
\begin{figure}
\centering
\begin{subfigure}{.5\textwidth}
  \centering
\includegraphics[clip, trim=6.1cm 8.5cm 6.1cm 1.5cm, width=\linewidth]{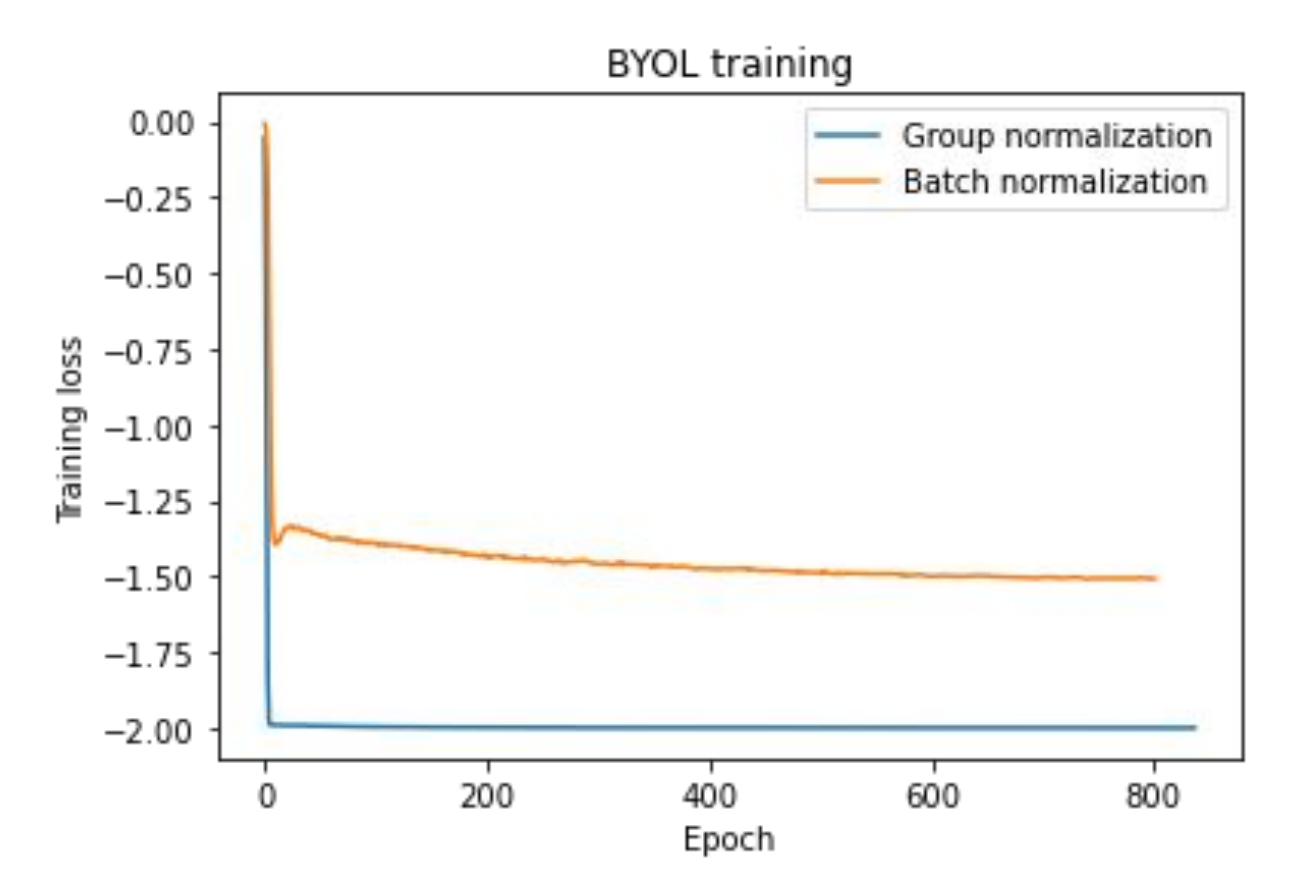}
    \caption{BYOL training}
    \label{fig:byol_bn_gn}
\end{subfigure}%
\begin{subfigure}{.5\textwidth}
  \centering
    \includegraphics[clip, trim=5.5cm 7.3cm 5.5cm 2cm, width=\linewidth]{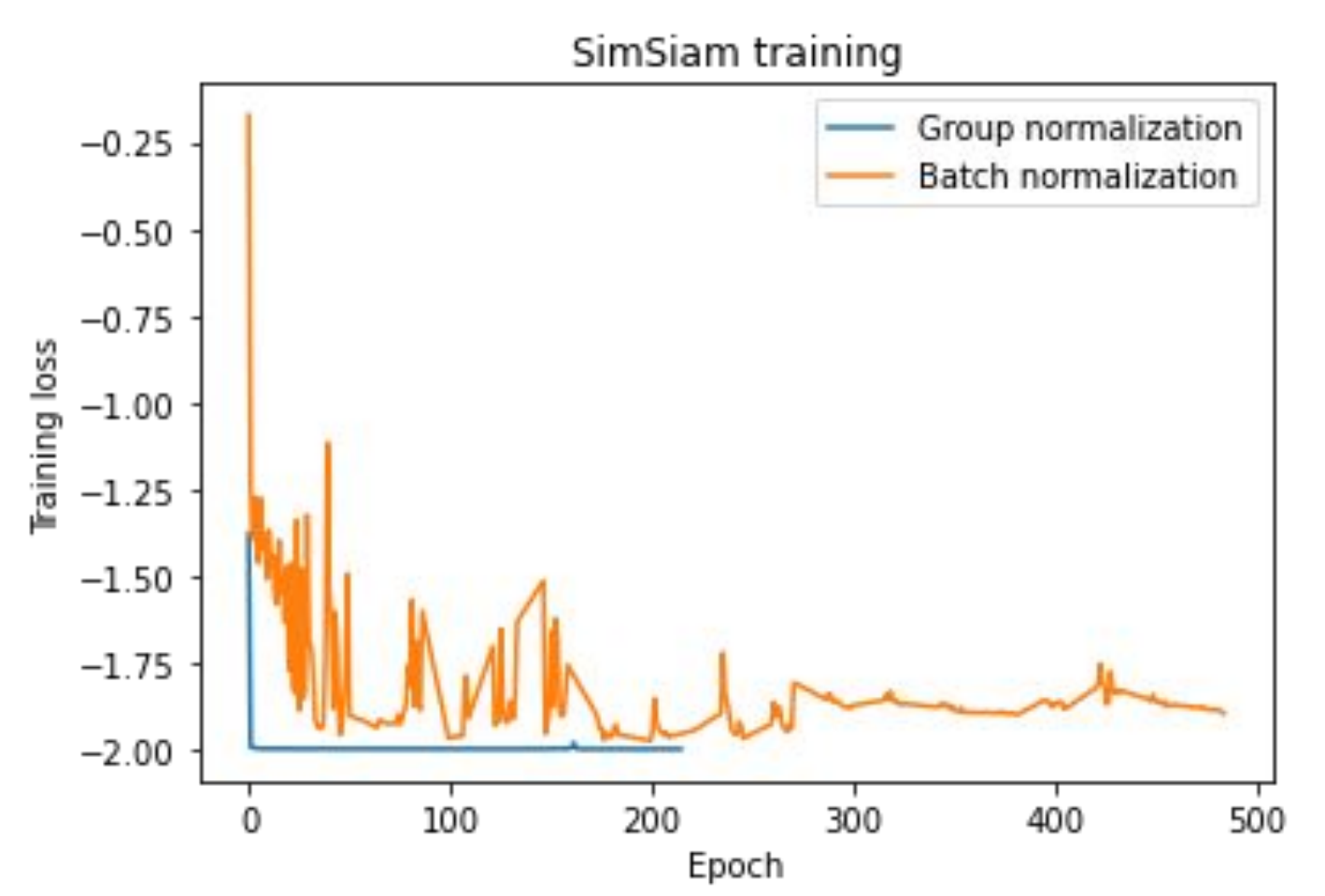}
    \caption{SimSiam training}
    \label{fig:simsiam_bn_gn}
\end{subfigure}
\caption{ResNet14 training on CIFAR-100 dataset.}
\label{fig:test}
\end{figure}

When we experimented with BYOL and SimSiam by replacing Batch Normalization (BN) with Group Normalization (GN), the models did not train well\footnote{~\citet{BN-BYOL} also observes that batch normalization is crucial for BYOL to work. Though ~\citet{GN-BYOL} refutes this and claims that BYOL works well with group normalization, we were not successful in our attempts.}. Figure~\ref{fig:byol_bn_gn} shows the training loss when ResNet14 encoder is trained with BYOL approach on CIFAR-100 dataset. When we use GN instead of BN, the loss quickly drops close to its lowest possible value and the model does not learn after that. When we evaluated the BN and GN-based BYOL encoders after training them for 800 epochs with a batch size of 512, we achieved 41\% and 3\% accuracy, respectively, under linear evaluation protocol. This clearly shows that BN is crucial for BYOL to work well. Similar behavior was observed in the case of SimSiam (see Fig.~\ref{fig:simsiam_bn_gn}).

\end{document}